\documentclass[10pt,twocolumn]{wlscirep}
\usepackage[utf8]{inputenc}
\usepackage[T1]{fontenc}
\usepackage{graphicx}
\usepackage{hyperref}
\usepackage[english]{babel}
\usepackage{rotating}
\usepackage{multirow}
\usepackage{booktabs}
\usepackage{bigstrut}
\usepackage{threeparttable}
\usepackage{siunitx} %
\DeclareSIUnit\pixel{px}
\DeclareSIUnit\bit{bit}
\DeclareSIUnit\byte{B}
\DeclareSIUnit[number-unit-product = ]\percent{\char`\%} %
\title{Search and Rescue with Airborne Optical Sectioning}

\author[*]{David C. Schedl}
\author[*]{Indrajit Kurmi}
\author[*]{Oliver Bimber}
\affil[*]{first.lastname@jku.at; Johannes Kepler University, Faculty of Engineering and Natural Sciences, Linz, 4040, Austria}

\begin{abstract}
We show that automated person detection under occlusion conditions can be significantly improved by combining multi-perspective images before classification. Here, we employed image integration by Airborne Optical Sectioning (AOS)---a synthetic aperture imaging technique that uses camera drones to capture unstructured thermal light fields---to achieve this with a precision/recall of \SI{96}{}/\SI{93}{\percent}. Finding lost or injured people in dense forests is not generally feasible with thermal recordings, but becomes practical with use of AOS integral images. Our findings lay the foundation for effective future search and rescue technologies that can be applied in combination with autonomous or manned aircraft. They can also be beneficial for other fields that currently suffer from inaccurate classification of partially occluded people, animals, or objects. 
\end{abstract}%
\begin{document}
\flushbottom
\maketitle
\thispagestyle{empty}

In 2018, \num{2723} search-and-rescue (SAR) incidents were reported by the National Park Service Units in \num{165} national parks throughout the United States. The operation costs added up to \SI{4.5}[\$]{M}, \SI{36}{\percent} of which were related to air operations. In the same year, \num{2292} alpine SAR operations were performed by \"OAMTC air emergency helicopters in Austria, and \num{2597} SAR missions were carried out by helicopters in the United Kingdom. In the UK \num{461} (\SI{18}{\percent}) of these flights searched for persons or crafts.

Rescuing, lost, ill or injured persons often involves searching densely forested terrain. Sunlight is mostly blocked by trees and other vegetation, and the forest ground reflects little light. Thermal imaging systems are therefore employed to visualize the temperature difference between human bodies and the surrounding environment. Autonomous unmanned drones will increasingly replace manned helicopters in future SAR operations,\cite{Burke2019,Lygouras2019} as they offer higher flexibility at lower cost. As in autonomous driving,\cite{Brunetti2018PedestrianDetectionSurvey,Yurtsever2020AutonomousDrivingSurvey} this requires for robust automatic people detection mechanisms. 

\begin{figure*}
\centering
\includegraphics[width=\linewidth]{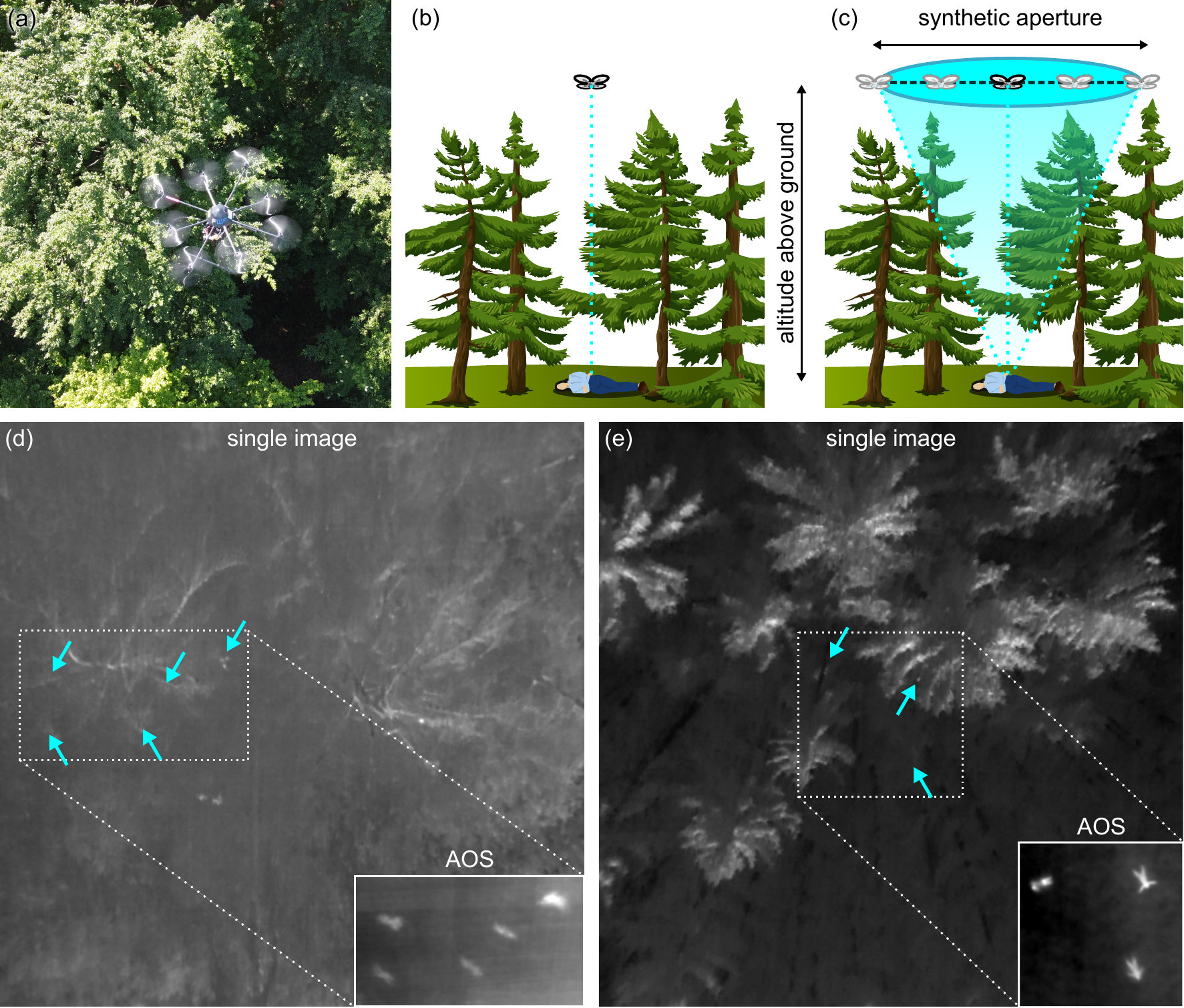}
\caption{ (a) Our drone autonomously scanning a forest patch. %
In contrast to recording and analyzing single images (b), AOS combines multiple images that are captured within a synthetic aperture before the resulting integral image is analyzed (c). %
Single thermal drone recordings at an altitude of \SI{35}{\m} above dense forest ground: (d) mixed forest, (e) conifer forest. The arrows indicate partial heat signals of occluded people on the ground. The insets show AOS results that are achieved when multiple thermal images are integrated. Note that contrast and brightness of the insets have been adjusted for better visibility. %
See Supplementary Video. %
}%
\label{fig:AOS} 
\end{figure*}

However, such search missions remain challenging due to occlusion and high heat radiation by trees under direct sunlight. Figure~\ref{fig:AOS} illustrates examples of thermal images of two different forest types (mixed and conifer forests)---captured from a drone---in which people on the ground can barely be detected because {\it (i)} their heat footprint is largely occluded by trees and {\it (ii)} the temperature of sunlight reflected by branches and tree crowns appears similar to body temperature on sensors. Obviously, simply thresholding the heat signal will not enable person detection.  

Synthetic apertures (SA) approximate the signal of a single hypothetical wide aperture sensor by means of either an array of static small aperture sensors or a single moving small aperture sensor whose individual signals are computationally combined to increase resolution, depth-of-field, frame rate, contrast, and signal-to-noise ratio. %
This principle has been used in fields such as radar,\cite{Moreira2013,Li2015,Rosen2000} radio telescopes,\cite{Levanda2010,Dravins2015} interferometric microscopy,\cite{Ralston2007} sonar,\cite{Hayes2009,Hansen2011} ultrasound,\cite{Jensen2006Review,Zhang2016} LiDAR,\cite{Barber2014,Turbide2017} and imaging.\cite{Vaish04,Vaish06,Zhang2018,YangTao2016,Joshi2007,Pei2019,YangTao2014,Pei2013}

With Airborne Optical Sectioning  (AOS)\cite{Kurmi2018,Bimberi2019IEEECGA,Kurmi2019TAOS,Kurmi2019Sensor,Schedl2020SRep}, we have introduced a synthetic-aperture imaging technique that uses a camera drone to capture an unstructured light field (i.e., a set of single images at unstructured sampling positions, Fig.~\ref{fig:AOS}; further details in Section~S1 of the Supplementary Material). Color and thermal images that are recorded within the shape of a wide synthetic aperture area above a forest are combined (registered and integrated) to computationally remove occluders, such as trees and other vegetation. Applied to thermal imaging, our technique makes  the radiated heat signal of largely occluded targets (e.g., a human body hidden in dense undergrowth) visible by integrating multiple thermal recordings from slightly different perspectives. The outcome is a mostly occlusion-free view of the forest ground. 

\begin{figure*}
\centering
\includegraphics[width=\linewidth]{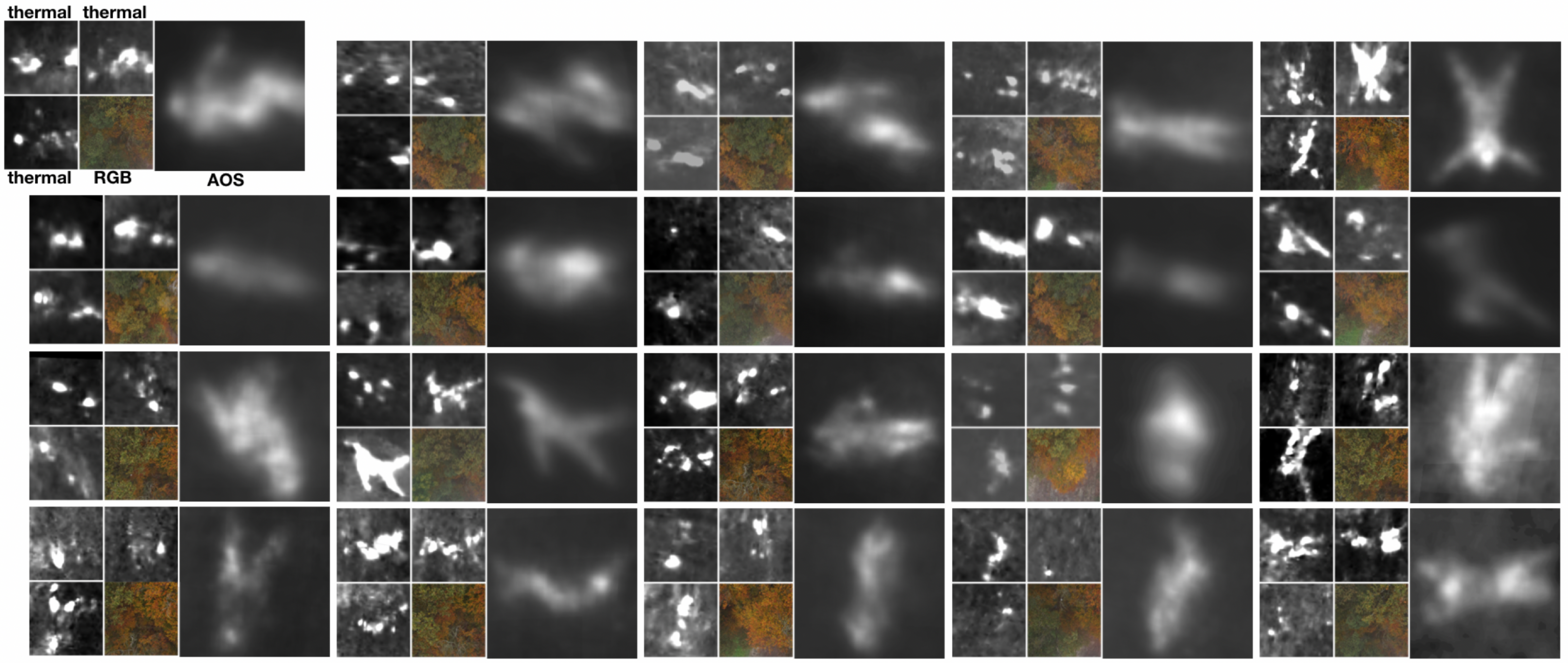}
\caption{Results of an initial field experiment. Twenty people lying on the ground in a dense broadleaf forest (zoom to the RGB samples for reference). For each person, a subset of 3 single thermal images (close-ups) and the corresponding AOS integral image (close-up) are shown. The \SI{30x30}{\m} synthetic aperture was scanned within \SI{10}{\minute} at an altitude of \SI{35}{\m} above ground.  %
}%
\label{fig:Initial_FE} 
\end{figure*}

Our hypothesis is that AOS integral images will enable hitherto unfeasible automated person detection in dense forests. Initial field experiments, such as that shown in Fig.~\ref{fig:Initial_FE}, corroborate our hypothesis. While the heat footprint in single thermal recordings shows mostly a random pattern due to varying partial occlusion from different views, AOS results often reveal the recognizable shape of a human. As mentioned above, considering only the strength of the heat signal is insufficient for detection, as similar heat signals are produced by direct reflection of sunlight from the trees.  

In this article we show that the detection rate can be significantly improved by combining multiple images (i.e., by registering and integrating) before detection rather than by combining multiple detection results from individual images. 
In our experiments, we achieve an average precision (AP) score of \SI{92.2}{\percent}, compared to an AP score of \SI{24.8}{\percent} with single images. %
Our findings pave the way for future autonomous SAR technologies that focus on finding lost and injured people in dense forests. 
Since fast operation is critical to such missions, computer-supported analysis of the enormous amount of image data is essential. However, human detection by means of AOS requires a specialized training dataset. Multi-spectral datasets that are available for autonomous driving,\cite{Hwang2015dataset,Xu2019dataset} for example, cannot be applied, as they mainly contain upright (i.e., standing, walking or running) people in urban environments and do not include AOS-specific optical aberrations (see Section~S2 in the Supplementary Material for an initial evaluation).

\begin{figure*}
\centering
\includegraphics[width=\linewidth]{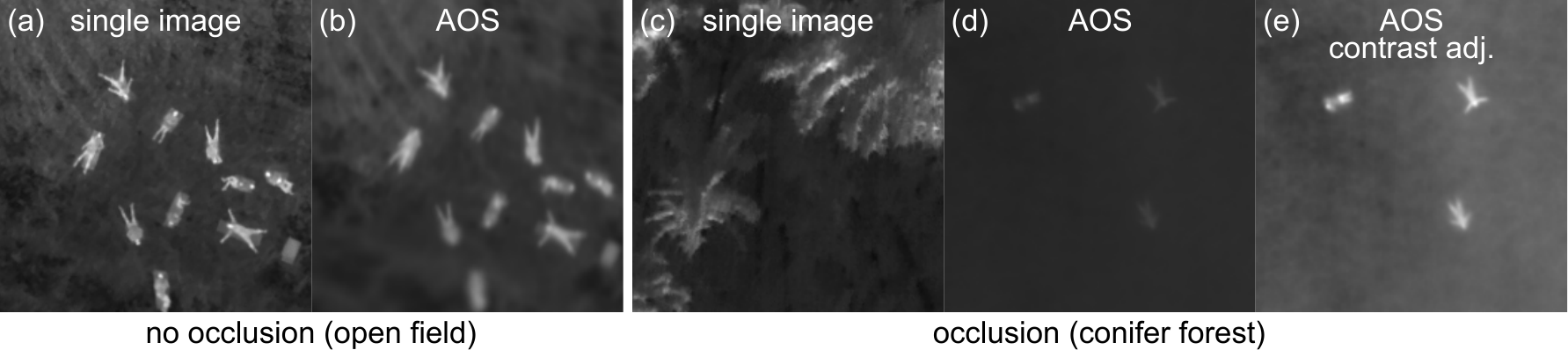}
\caption{Comparison of scenes without (a--b) and with (c--e) occlusion in terms of contrast level. The differences between AOS integral images without (b) and with (d) occlusion are small after contrast adjustment (e). %
}%
\label{fig:contrast} 
\end{figure*}

Results of our initial field experiments (cf. Fig.~\ref{fig:contrast}) show that occlusion has little effect on the AOS image-forming results. The more occlusion, the lower the contrast in AOS integral images, as explained by the statistical model presented in previous work.\cite{Kurmi2019Sensor} %
After contrast adjustment, however, human shapes and optical aberrations (defocus) can be identified, regardless of whether occlusion was present. This indicates that a training dataset can be produced under controlled open-field conditions (i.e., without occlusion) rather than in forests of different types and densities. 

\section*{Results} \label{sec:results}
Test and training data for our experiments were recorded in 18 drone flights (see Table~\ref{tab:dataset}; Section~S3 in the Supplementary Material) in close proximity to Linz, Austria. 
Twelve flights were performed above forests of various vegetation types (broad-leaved, conifer and mixed forests) at an altitude of approximately \SIrange{30}{35}{\m} above ground layer (AGL). The remaining 6 flights recorded data from above a meadow without any high vegetation. To protect subjects in this open field, a safety net (\SI{2}{\m} AGL) was installed. The net strings are not resolvable in integral images captured from the recording altitude. 

Subjects were asked to lie on the ground (each in a random pose) and remain still or perform little motion (such as waving hands) to simulate ill or injured persons. Flights covered a square synthetic aperture area of \SI{30 x 30}{\m} with \SI{1 x 3}{\m} dense sampling (exceptions are indicated in Table~\ref{tab:dataset}). %
A low-resolution (\SI{640 x 512}{\pixel}) thermal and a high-resolution (\SI{6000 x 4000}{\pixel}) RGB camera were triggered simultaneously while the drone was flying at a speed of \SI{0.7}{\meter\per\second}, capturing approximately \num{360} thermal and RGB image pairs in the course of approximately \SI{10}{\minute} flights. 

After recording, the high-resolution RGB images were used for precise pose estimation. 
Since the drone's altitude above ground was known, approximate focus parameters could be pre-estimated (for larger terrain, digital elevation models can be used). Minor variations in ground elevation were handled by a local parameter optimization.\cite{Kurmi2020} %
Thus, occluding vegetation was suppressed and focused humans on the ground were emphasized in thermal integral images, as illustrated in Fig.~\ref{fig:AOS}. 
In each integral image, people were manually labeled by polygonal contours. Since camera poses and focus parameters are known, the polygonal contour points are three-dimensional and can be related to other camera poses or focus settings. The number of labels (persons) for each flight is shown in Table~\ref{tab:dataset}.

\begin{table*}[htbp]  \centering
\begin{threeparttable}[b]
  \caption{Experimental dataset (see Fig.~S3 for exemplary RGB and thermal images). Latitude angles are north of the equator and longitude angles are east of the prime meridian.}
\begin{tabular}{lllcrccc}
\textbf{ID} & \textbf{Latitude} & \textbf{Longitude} & \textbf{Forest} & \multicolumn{1}{c}{\textbf{Date}} & \multicolumn{1}{c}{\textbf{Labels}} & \textbf{Set} \\
F0\tnote{a} & \ang{48;25;17.32} & \ang{14;18;10.08} & conifer & 4 Oct 19 & 3     & test \\
F1\tnote{b} & \ang{48;19;59.42} & \ang{14;19;52.40} & broadleaf & 24 Oct 19 & 10    & test \\
F2    & \ang{48;19;59.56} & \ang{14;19;52.77} & broadleaf & 24 Oct 19 & 10    & test \\
F3    & \ang[parse-numbers=false]{48;20;01.41} & \ang{14;19;48.50} & mixed & 25 Oct 19 & 6     & test \\
F4    & \ang[parse-numbers=false]{48;20;01.56} & \ang{14;19;48.74} & mixed & 25 Oct 19 & 6     & test \\
F5    & \ang{48;19;57.70} & \ang{14;19;48.35} & conifer & 8 Nov 19 & 10    & test \\
F6    & \ang{48;19;57.70} & \ang{14;19;48.35} & conifer & 8 Nov 19 & 10    & test \\
F7\tnote{c} & \ang{48;19;59.56} & \ang{14;19;52.18} & broadleaf & 20 Nov 19 & 2     & test \\
O1    & \ang{48;20;16.46} & \ang{14;18;50.52} & none  & 8 Jan 20 & 10    & train \\
O2    & \ang{48;20;16.46} & \ang{14;18;50.52} & none  & 8 Jan 20 & 10    & train \\
F8    & \ang{48;19;59.19} & \ang{14;19;52.11} & broadleaf & 17 Jan 20 & 0     & train \\
F9    & \ang{48;19;58.66} & \ang{14;19;51.71} & broadleaf & 17 Jan 20 & 0     & test \\
O3    & \ang{48;20;16.46} & \ang{14;18;50.52} & none  & 22 Jan 20 & 6     & train \\
O4    & \ang{48;20;16.46} & \ang{14;18;50.52} & none  & 22 Jan 20 & 6     & train \\
O5    & \ang{48;20;16.46} & \ang{14;18;50.52} & none  & 7 Feb 20 & 5     & train \\
O6    & \ang{48;20;16.46} & \ang{14;18;50.52} & none  & 7 Feb 20 & 5     & valid \\
F10   & \ang[parse-numbers=false]{48;20;01.75} & \ang{14;19;48.92} & mixed & 10 Apr 20 & 0     & train \\
F11   & \ang{48;19;57.60} & \ang{14;19;48.39} & conifer & 10 Apr 20 & 0     & test \\
\end{tabular}%
    \begin{tablenotes}
    \item [a] \SI{1 x 2}{\m} spacing, resulting in \num{402} images. 
    \item [b] aborted early and contains only \num{153} images.  
    \item [c] \SI{5}{\m} circular synthetic aperture with \num{31} images. 
    \end{tablenotes}
     \label{tab:dataset}%
   \end{threeparttable}
\end{table*}%

The recorded data was split into training, validation and test sets. %
As previously mentioned, our initial experiments verified that human shapes and optical aberrations (defocus) can be identified in integral images---both with and without occlusion. Thus, for training we used 5 open-field scenes with 37 labels, 2 empty forest scenes, and 1 open-field scene with 5 labels for validation. 

For classifying people in our data, we trained the You Only Look Once (YOLO) \cite{Redmon2016yolov1,Redmon2017yolov2,Redmon2018yolov3} deep learning object detector, which supports fast detection rates, can run on embedded low-cost and low-power systems\cite{Shafiee2017,Vandersteegen2019,Yang2020} (e.g., as used on drones), and has proved its applicability to thermal object detection tasks in previous studies.\cite{Campilho2018,Ivavsic2019} 

To increase the number of samples in the training and validation sets, we used common data augmentation techniques. For instance, we optionally applied adaptive histogram equalization (AHE) to every integral image and added the resulting images to the dataset. 
AHE reduces the effect of temperature variations and has been used previously to enhance thermal images.\cite{Zheng2019} %
The orientation of the images was also changed randomly 10 times to account for various rotations of the test subjects. Furthermore, we altered the focus parameters, because multiple heat sources on the ground or non-planar ground may lead to slight defocus for a single focal plane setting. 
We applied 27 focus variations by translating focus away and towards the ground and by rotating the focal plane about its two axes. 

Since our labels are 3D polygonal contours, they can be converted to correct axis-aligned bounding boxes after augmentation. Ultimately, augmentation resulted in a total of \num{540} images with corresponding labels for every flight. %
Further augmentations were performed during training by the training algorithm of YOLO and include random horizontal image flipping and minor brightness changes.
We used the average precision (AP) metric\cite{VOC2011} for the validation dataset to determine when to stop training.

After training, we applied the resulting network to the test dataset. 
At test time, we optionally performed AHE, ran the detection twice (on both the non-augmented and the augmented images), and combined the two results. To avoid potential double detections, we applied non-maximum suppression (NMS).   

\begin{figure*}
\centering
\includegraphics[width=\linewidth]{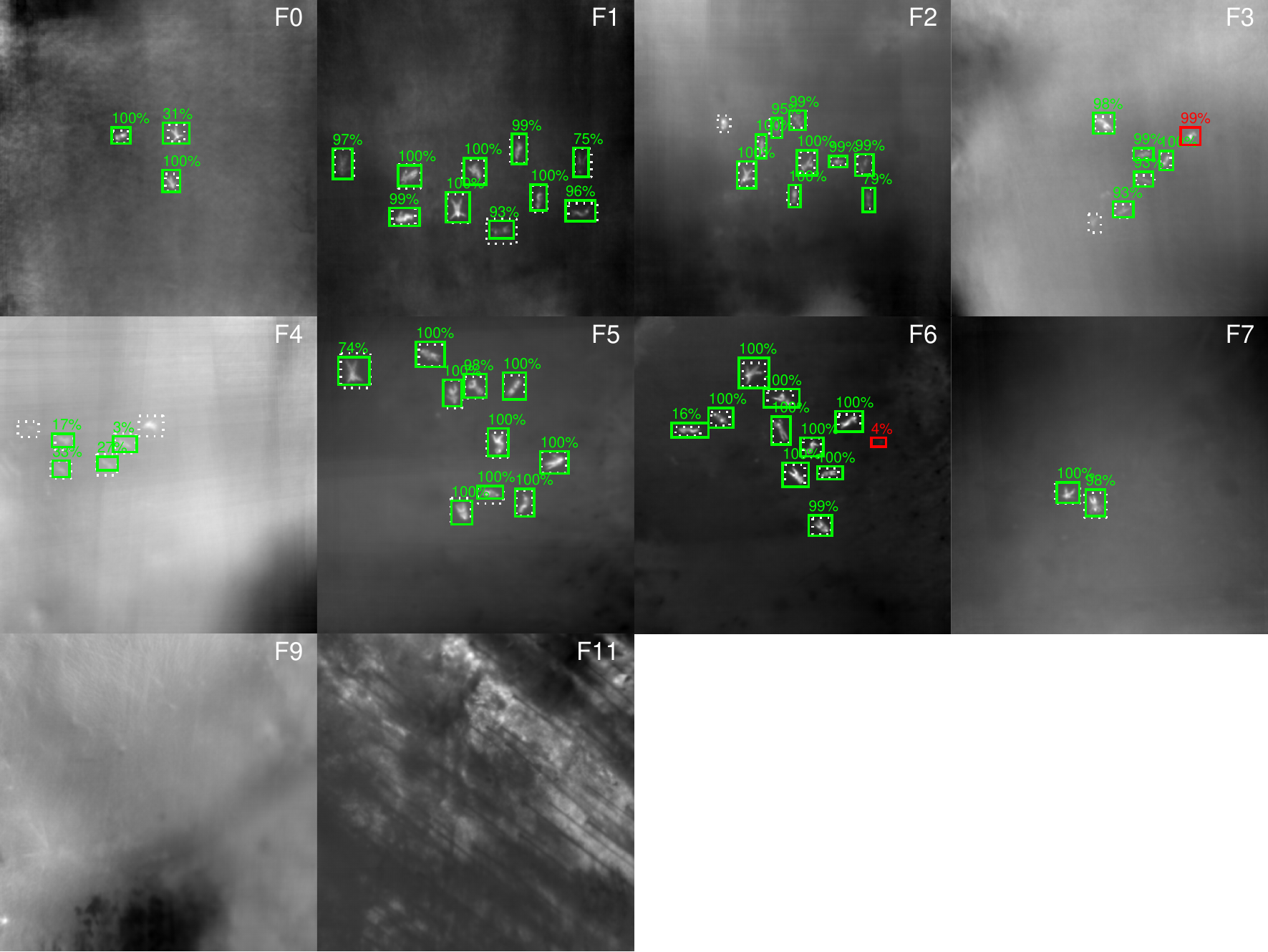}
\caption{AOS person detection results for the 10 test scenes (cf. Table~\ref{tab:dataset}, AHE augmentation applied to both test and training sets). The ground truth labels are enclosed within white dashed rectangles. Detections are indicated by solid rectangles, where TPs are green and FPs are red. The numbers above bounding boxes indicate the network's confidence score. Corresponding AP scores and the numbers of TPs and FPs are shown in Table~\ref{tab:results_AOS} (column "AHE train + test set"). Note that F9 and F11 are empty.%
}%
\label{fig:results_AOS} 
\end{figure*}

For AHE applied to both training and test set, we achieved an overall AP score of \SI{92.2}{\percent} (precision/recall \SI{96.4}{}/\SI{93}{\percent}; IoU threshold \SI{25}{\percent}) and detected \num{53} out of \num{57} persons (true positives) and only \num{2} false positives (branches and a dog, were classified as persons) in the test dataset. %
Visual results are shown in Fig.~\ref{fig:results_AOS}.

\begin{figure*}
\centering
\includegraphics[width=\linewidth]{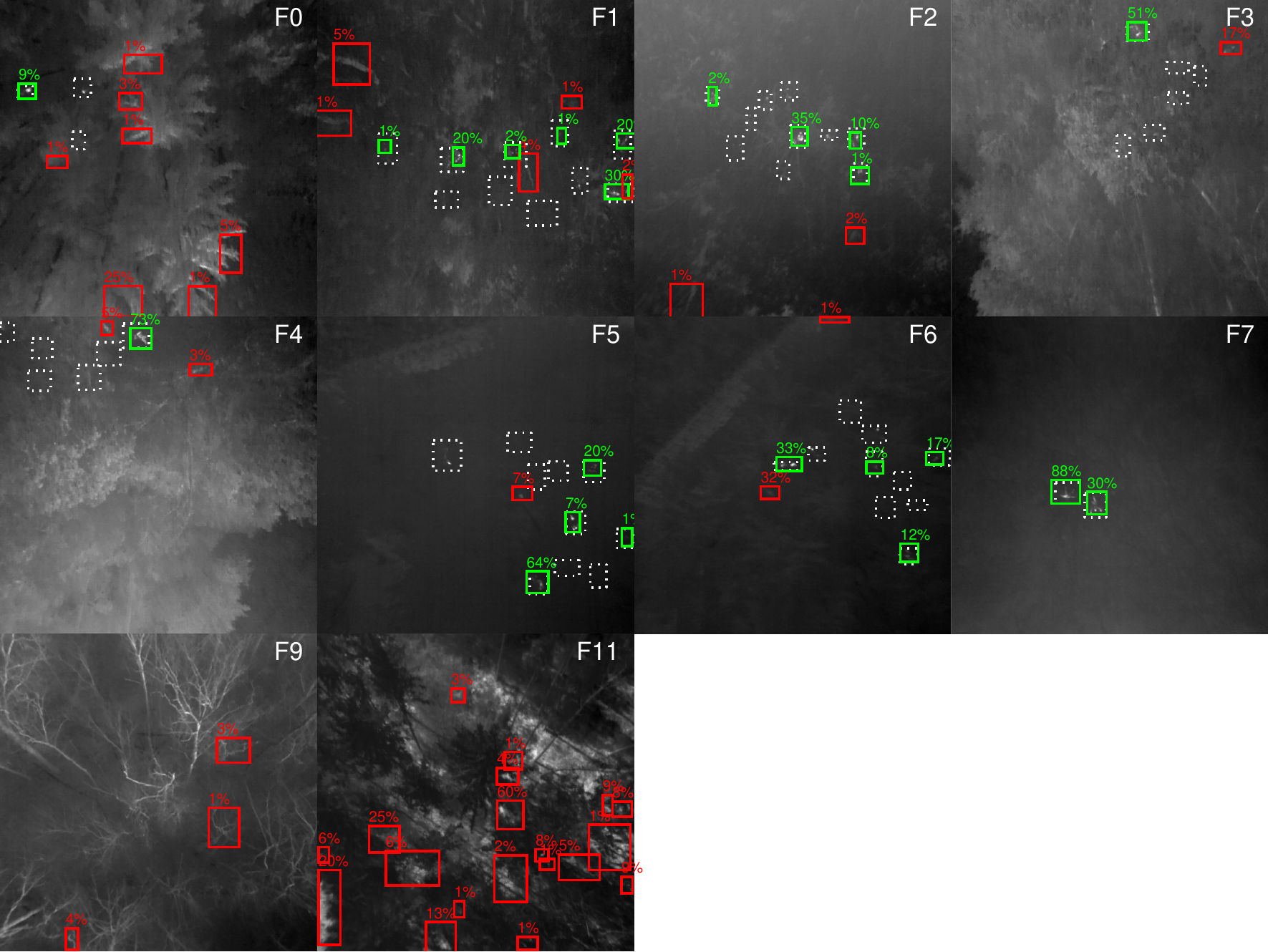}
\caption{Single-image person detection results for the 10 test scenes (cf. Table~\ref{tab:dataset}, AHE augmentations applied to test set only). The ground truth labels are enclosed within dashed rectangles. Detections are indicated by solid rectangles, where TPs are green and FPs are red. The numbers are the network's confidence scores. Corresponding AP scores and the numbers of TPs and FPs are shown in Table~\ref{tab:results_singleImg} (column "AHE test set only").%
}%
\label{fig:results_singleImg} 
\end{figure*}

We compared AOS detection to a conventional single-image detection in terms of performance, and classified all single images of the test set. For this purpose, a new network was trained with the single images of the training set. Three-dimensional labels were transferred automatically from the integral images to the single images. The augmentations used for the integral images were reused, with the exception of the focus parameters (which do not apply to single images). After training, the network was applied to the single images of the test set.
As indicated by our initial experiments, the single-image detection rate dropped significantly to a maximum AP score of \SI{24.8}{\percent} (precision/recall scores \SI{25.1}{}/\SI{50.4}{\percent}) when AHE was applied to the test set only (cf. Fig.~\ref{fig:results_singleImg}). 

Tables~\ref{tab:results_AOS} and \ref{tab:results_singleImg} show AP scores, ground truth (GT), and the number of true positives (TP) and false positives (FP) for our test scenes (rows) when using AOS and single images, respectively. The tables list the detection results for the networks trained with and without AHE. For testing, results with and without AHE applied in combination with NMS are listed. 

As shown in Table~\ref{tab:results_AOS}, applying AHE to the training and test datasets clearly resulted in the best detection performance for AOS integral images. %
For single images, AHE reduced FP when applied to the test set (i.e., the average FP rate dropped from \num{4.5} to \num{0.7} without and from \num{5.4} to \num{1.5} with AHE test augmentation), but also reduced the TP rate by a factor of approximately \num{2}. %
This drop explains the lower AP scores of the results with training augmentation compared to those without, as summarized in Table~\ref{tab:results_singleImg}. %

\begin{table*}[htbp]
  \centering
  \caption{AOS person detection results. Average precision scores (AP), ground truth (GT), true positives (TP), and false positives (FP) for the integral image of each scene. Augmentation with AHE: not applied, applied to test set only, applied to training set only, applied to test and training set.}
\begin{tabular}{r@{ }l|rrr|rrr|rrr|rrr}%
      &       & \multicolumn{3}{c|}{\textbf{no AHE}} & \multicolumn{3}{c|}{\textbf{AHE test set only}} & \multicolumn{3}{c|}{\textbf{AHE train set only}} & \multicolumn{3}{c}{\textbf{AHE train + test set}} \\
\textbf{ID} & \textbf{(GT)} & \multicolumn{1}{c}{\textbf{AP}} & \multicolumn{1}{l}{\textbf{FP}} & \multicolumn{1}{l|}{\textbf{TP}} & \multicolumn{1}{c}{\textbf{AP}} & \multicolumn{1}{l}{\textbf{FP}} & \multicolumn{1}{l|}{\textbf{TP}} & \multicolumn{1}{c}{\textbf{AP}} & \multicolumn{1}{l}{\textbf{FP}} & \multicolumn{1}{l|}{\textbf{TP}} & \multicolumn{1}{c}{\textbf{AP}} & \multicolumn{1}{l}{\textbf{FP}} & \multicolumn{1}{l}{\textbf{TP}} \\
\hline
F0    & (3)   & \textbf{100.0\%} & 0     & \textbf{3} & \textbf{100.0\%} & 0     & \textbf{3} & \textbf{100.0\%} & 0     & \textbf{3} & \textbf{100.0\%} & 0     & \textbf{3} \\
F1    & (10)  & \textbf{100.0\%} & 0     & \textbf{10} & \textbf{100.0\%} & 0     & \textbf{10} & 60.0\% & 0     & 6     & \textbf{100.0\%} & 0     & \textbf{10} \\
F2    & (10)  & 80.0\% & 0     & 8     & \textbf{90.0\%} & 0     & \textbf{9} & 30.0\% & 0     & 3     & \textbf{90.0\%} & 0     & \textbf{9} \\
F3    & (6)   & 16.7\% & 0     & 1     & 50.0\% & 0     & 3     & 16.7\% & 0     & 1     & \textbf{73.1\%} & 1     & \textbf{5} \\
F4    & (6)   & 0.0\% & 0     & 0     & 16.7\% & 0     & 1     & 0.0\% & 0     & 0     & \textbf{66.7\%} & 0     & \textbf{4} \\
F5    & (10)  & \textbf{100.0\%} & 0     & \textbf{10} & \textbf{100.0\%} & 0     & \textbf{10} & 90.0\% & 0     & 9     & \textbf{100.0\%} & 0     & \textbf{10} \\
F6    & (10)  & \textbf{100.0\%} & 1     & \textbf{10} & 99.1\% & 2     & \textbf{10} & \textbf{100.0\%} & 0     & \textbf{10} & \textbf{100.0\%} & 1     & \textbf{10} \\
F7    & (2)   & \textbf{100.0\%} & 0     & \textbf{2} & \textbf{100.0\%} & 1     & \textbf{2} & 50.0\% & 0     & 1     & \textbf{100.0\%} & 0     & \textbf{2} \\
F9    & (0)   & \multicolumn{1}{c}{n/a} & 0     & 0     & \multicolumn{1}{c}{n/a} & 0     & 0     & \multicolumn{1}{c}{n/a} & 0     & 0     & \multicolumn{1}{c}{n/a} & 0     & 0 \\
F11   & (0)   & \multicolumn{1}{c}{n/a} & 0     & 0     & \multicolumn{1}{c}{n/a} & 0     & 0     & \multicolumn{1}{c}{n/a} & 0     & 0     & \multicolumn{1}{c}{n/a} & 0     & 0 \bigstrut[b]\\
\hline
\textbf{sum} & (57)  & 77.2\% & 1     & 44    & 83.2\% & 3     & 48    & 57.9\% & 0     & 33    & \textbf{92.2\%} & 2     & \textbf{53} \bigstrut[t]\\
\end{tabular}%
  \label{tab:results_AOS}%
\end{table*}%
\begin{table*}[htbp]
  \centering
  \caption{Single-image person detection results. Average precision scores (AP), ground truth (GT), true positives (TP), and false positives (FP) as averages over all images of each scene. Augmentation with AHE: not applied, applied to test set only, applied to training set only, applied to test and training set.}
    \begin{tabular}{r@{ }l|rrr|rrr|rrr|rrr}%
      &       & \multicolumn{3}{c|}{\textbf{no AHE}} & \multicolumn{3}{c|}{\textbf{AHE test set only}} & \multicolumn{3}{c|}{\textbf{AHE train set only}} & \multicolumn{3}{c}{\textbf{AHE train + test set}} \\
\textbf{ID} & \textbf{(avg. GT)} & \multicolumn{1}{c}{\textbf{AP}} & \multicolumn{1}{l}{\textbf{FP}} & \multicolumn{1}{l|}{\textbf{TP}} & \multicolumn{1}{c}{\textbf{AP}} & \multicolumn{1}{l}{\textbf{FP}} & \multicolumn{1}{l|}{\textbf{TP}} & \multicolumn{1}{c}{\textbf{AP}} & \multicolumn{1}{l}{\textbf{FP}} & \multicolumn{1}{l|}{\textbf{TP}} & \multicolumn{1}{c}{\textbf{AP}} & \multicolumn{1}{l}{\textbf{FP}} & \multicolumn{1}{l}{\textbf{TP}} \bigstrut[b]\\
\hline
F0    & (2.6) & 4.0\% & 7.1   & \multicolumn{1}{r|}{0.8} & \textbf{7.6\%} & 8.8   & \textbf{1.0} & 4.3\% & 1.4   & \multicolumn{1}{r|}{0.4} & 5.0\% & 1.6   & 0.4 \bigstrut[t]\\
F1    & (7.2) & 55.5\% & 4.3   & \multicolumn{1}{r|}{4.6} & \textbf{57.1\%} & 4.7   & \textbf{4.7} & 27.5\% & 0.7   & \multicolumn{1}{r|}{2.1} & 28.1\% & 0.8   & 2.1 \\
F2    & (8.5) & 24.5\% & 2.6   & \multicolumn{1}{r|}{2.8} & \textbf{33.7\%} & 3.4   & \textbf{3.7} & 11.6\% & 0.1   & \multicolumn{1}{r|}{1.0} & 11.7\% & 0.2   & 1.0 \\
F3    & (4.4) & 14.7\% & 0.7   & \multicolumn{1}{r|}{0.8} & \textbf{20.9\%} & 1.4   & \textbf{1.2} & 3.2\% & 0.0   & \multicolumn{1}{r|}{0.2} & 3.1\% & 0.1   & 0.2 \\
F4    & (2.6) & 3.4\% & 1.2   & \multicolumn{1}{r|}{0.3} & \textbf{7.5\%} & 2.3   & \textbf{0.5} & 1.9\% & 0.1   & \multicolumn{1}{r|}{0.1} & 1.8\% & 0.2   & 0.1 \\
F5    & (5.8) & 57.3\% & 1.0   & \multicolumn{1}{r|}{3.5} & \textbf{63.5\%} & 1.2   & \textbf{3.8} & 34.5\% & 0.3   & \multicolumn{1}{r|}{2.1} & 35.0\% & 0.4   & 2.1 \\
F6    & (5.7) & 72.6\% & 2.0   & \multicolumn{1}{r|}{4.3} & \textbf{75.7\%} & 2.2   & \textbf{4.4} & 55.2\% & 0.7   & \multicolumn{1}{r|}{3.2} & 55.9\% & 0.9   & 3.2 \\
F7    & (2.0) & 75.4\% & 0.1   & \multicolumn{1}{r|}{1.5} & \textbf{96.6\%} & 0.2   & \textbf{1.9} & 51.5\% & 0.0   & \multicolumn{1}{r|}{1.0} & 51.5\% & 0.0   & 1.0 \\
F9    & (0.0) & \multicolumn{1}{c}{n/a} & 2.1   & \multicolumn{1}{r|}{0.0} & \multicolumn{1}{c}{n/a} & 2.9   & 0.0   & \multicolumn{1}{c}{n/a} & 1.6   & \multicolumn{1}{r|}{0.0} & \multicolumn{1}{c}{n/a} & 4.5   & 0.0 \\
F11   & (0.0) & \multicolumn{1}{c}{n/a} & 16.5  & \multicolumn{1}{r|}{0.0} & \multicolumn{1}{c}{n/a} & 18.0  & 0.0   & \multicolumn{1}{c}{n/a} & 0.8   & \multicolumn{1}{r|}{0.0} & \multicolumn{1}{c}{n/a} & 4.8   & 0.0 \bigstrut[b]\\
\hline
\textbf{avg} & (3.6) & 18.0\% & 4.5   & 1.6   & \textbf{24.8\%} & 5.4   & \textbf{1.8} & 19.0\% & 0.7   & 0.8   & 17.7\% & 1.6   & 0.9 \bigstrut[t]\\
\end{tabular}%
  \label{tab:results_singleImg}%
\end{table*}%

\section*{Conclusion and future work} 
We have shown that the detection rate for human classification under occlusion conditions can be significantly increased by combining multi-perspective recordings before classification. While commonly used real-time classifiers (such as YOLO) produce poor results on single images, they produce usable outputs on integral images. We have also demonstrated that the training data for our approach is widely invariant to occlusion, and can therefore be generated easily under controlled (open-field) conditions.

We believe, that these findings provide a foundation for future autonomous SAR technologies that focus on finding lost and injured people in dense forests. Our approach can also assist conventional SAR missions that are carried out with manned helicopters or airplanes. Furthermore, it could  support the surveillance of humans in the course of military and law-enforcement tasks, the monitoring of animals for wildlife observation, or  autonomous vehicles whenever person classification suffers from occlusion. However, with a world-wide increase of drone applications, new challenges, concerning ethics, sustainability, security, and privacy, arise and need to be addressed.\cite{Finn2012,Rao2016,Shakhatreh2019} Camera drones should adhere to privacy regulations and respect human rights. 

Registration and averaging is just one possible way of combining images from multi-perspective recordings with AOS, and we are planning to explore further options. %
For instance, AOS also supports computing  entire focal stacks (i.e., integral images for multiple, axially varying synthetic focal planes), which better preserve depth information than single integral images and might increase detection rates further. 
This approach requires a network structure that operates on volumetric data,\cite{Lu2019volume} which we will investigate as part of future work. 

Our current implementation could clearly be improved by more sophisticated detection techniques \cite{Tan2020EfficientDet,Zhang2020ATSS,Liu2019ASFF,Lee2019CenterMask} and larger training sets with more human participants.  
Advanced data augmentation techniques might also lead to higher detection rates. %
First experiments applying augmentation techniques that simulate occlusions in single images, however, have not improved the detection performance and are described in Section~S4 of the Supplementary Material.
Notwithstanding these considerations, we believe that combining multiple perspectives before classification will continue to produce superior results. 
One of our biggest limitation is the short battery life of camera drones that restricts flight time and thus the scanning coverage. Therefore, we are currently investigating the efficiency of one-dimensional line scans (i.e., 1D synthetic apertures) for person detection, rather than two-dimensional area scans. Initial experiments indicate that 1D apertures are sufficient and allow to cover a significantly larger range. 
Furthermore, we are working on a first fully embedded on-board implementation that carries out all measurements and computations directly on the drone and during flight. It will support real-time rates (so far, approximately \SI{200}{\milli\second} are needed for image integration  and classification using a Raspberry Pi and an Intel Neural Compute Stick that runs YOLO-tiny\cite{Redmon2018yolov3} (a YOLO version optimized for mobile processors).
\section*{Methods} %
We recorded our datasets using an octocopter (MikroKopter OktoXL 6S12; \SI{945}{\milli\meter} diameter; approx.~\SI{4.9}{\kg}; two LiPo \SI{4500}{\milli\ampere\hour} batteries), equipped with a thermal camera (Flir Vue Pro; \SI{9}{\mm} fixed focal length lens; \SIrange{7.5}{13.5}{\micro\meter} spectral band; \SI{14}{\bit} non-radiometric) and an RGB camera (Sony Alpha 6000; \SIrange{16}{50}{\mm} lens at infinite focus).
The cameras were fixed to a rotatable gimbal, were triggered synchronously (synched by a MikroKopter CamCtrl control board), and pointed downwards during all flights.
A synthetic aperture of \SI{30x30}{\meter} was chosen, because at an altitude of \SI{30}{\m} (maximal tree height plus safety margin) the field of view of all recorded single images is just overlapping on the ground. %
The aperture's flight pattern was planned using MikroKopter's flight planning software and uploaded to the drone as waypoints. The waypoint protocol triggered the cameras every \SI{1}{\m} along the flight path, and the recorded images were stored on the cameras' internal memory cards.

After landing the drone and downloading the images from the memory cards, we processed the recorded data on a personal computer. %
To estimate the drone's pose, we used the unprocessed RGB images (\SI{6000 x 4000}{\pixel}) together with the general-purpose structure-from-motion and multi-view stereo pipeline, COLMAP.\cite{schonberger2016sfm} %
COLMAP required approximately 24 minutes for pose estimations of 300 images in our implementation. 
Since the cameras were fixed to a gimbal, the poses of the thermal camera could be directly obtained from the poses of the RGB camera by means of a precalibrated transformation matrix, which was computed using Matlab's checkerboard calibration routine. The calibration checkerboard was made of metal with black velvet checkers and is detectable in both thermal and RGB images.\cite{Kurmi2019TAOS} %
The thermal images were rectified to remove lens distortions and cropped to a field of view of \ang{50.82} and a resolution of \SI{512 x 512}{\pixel}. %
For rectification we applied OpenCV’s pinhole camera model.\cite{Zhang2000CameraCalibration} %
Since our thermal camera was non-radiometric, sensor readings did not correspond to absolute (but to relative) temperatures and changed continuously. Therefore, the thermal images' intensity mean was adjusted to the same range for each recorded scene. 

The integral images were computed on a GPU using our visualization technique\cite{Kurmi2018} implemented with Nvidia's CUDA toolkit. %
The integration of \num{360} single images took \SI{60}{\milli\second} on our system. %
For integral image visualization, a virtual camera was placed within the synthetic aperture's center, and its field of view was set to \ang{50.82} (i.e., the single-image field of view after rectification). Optimal settings for the synthetic focal plane were obtained by optimization.\cite{Kurmi2020} %
Note that the automatic focal plane optimization did not focus on the ground of scenes F8 and F9, as there were no distinguishable heat sources. Thus, focal plane adjustment was done manually in these two cases. More details on generating AOS integral images can be found in our previous publications\cite{Kurmi2018,Kurmi2019TAOS} and in Section~S1 of the Supplementary Material.  

Persons were labeled in single (non-augmented) integral images using Matlab's polygonal tool. %
The labels were converted to axis-aligned bounding boxes (AABBs) after augmentation (i.e., rotation) and stored as text files at the location of the corresponding images. %
Although the classifier as well as AP, TP and FP computations require AABBs, we used polygonal labels as intermediate representations, since they remain unaffected by rotation augmentation. %

For the test and the validation datasets (see Table~\ref{tab:dataset}), we applied the following augmentation techniques to the integral images: We randomly rotated the images by changing the direction of the virtual camera's up-vector in our visualization (corresponding to a rotation about the image center). %
Furthermore, we modified the focal plane parameters. 
We changed the altitude of the focal plane by $\pm$ \SI{0.25}{\m} around the optimal focal plane, and rotated it about its vertical and horizontal axes (relative to the virtual camera's orientation) by $\pm$ \SI{2}{\degree}. 
This led to a fixed number of \num{270} augmentations per scene. %
Note that our augmentation pipeline operated on single-precision floating-point (\SI{32}{\bit}) high-dynamic-range (HDR) images. 
The optional adaptive histogram equalization was performed using Matlab's Image Processing Toolbox after all other augmentation steps had been applied. For HDR images, the number of histogram bins was increased to \num{512} while all other parameters were kept at their default settings. Finally, the HDR images were tone-mapped to \SI[number-unit-product={\text{-}}]{8}{\bit} LDR grayscale images (required by the classifier) and normalized within the range of the hottest and coldest relative temperature readings per image. 

Augmentations of single images (i.e., rotation and optional AHE) were performed in Matlab, as explained above. 
Blank borders that were introduced when rotating rectangular images were removed by cropping. %
The labels of the integral images were directly projected to the single images using the known pose matrices and focal plane parameters (the contours define a 3D plane). 
Projected labels at borders were truncated and discarded if their clipped AABBs were less than \SI{25}{\percent} of their previous sizes. After label transfer, all images were inspected manually and outliers (due to, e.g., strong pose estimation errors) were removed. 

We trained YOLO\cite{Redmon2018yolov3} version 3 as implemented in the Darknet framework,\cite{Alexey2020darknet,Bochkovskiy2020yolov4} employing the network structure that utilizes spatial pyramid pooling (SPP).\cite{He2015spp,Huang2020spp} %
The default YOLOv3 SPP configuration with \num{114} layers was used, and only the changes required to support single-class (person) detection were applied.
The input image size of the network was set to \SI{512 x 512}{\pixel} and matched the resolution of the integral images and the rectified (non-cropped) single images. %
Training was performed on two NVIDIA GTX 2070 GPUs, and the network's batch and subdivision sizes were set to \num{64} to fit into \SI{8}{\giga\byte} of GPU RAM. %
During training, YOLOv3 performed further augmentations internally, including random horizontal image flipping and minor brightness changes. Augmentations unsuitable for grayscale images (e.g., hue or saturation changes) were turned off.
For training we used \num{53} convolutional filters that were pre-trained on Imagenet\cite{Russakovsky2015ImageNet} for the first \num{75} network layers (the backbone). %
The starting learning rate was set to \num{0.001}, and training weights were stored after every \num{200} batch iterations. 
After training, the stored training weights were used to compute AP scores (with IoU \SI{50}{\percent} using Darknet) on the validation datasets, and the weight with the highest AP result was used as final weight. Note that for the validation and training sets the same training augmentations were applied. Test augmentations were not applied to the validation datasets. The optimal weights were obtained after \SIrange{3200}{4600}{} iterations or \SIrange{4}{143}{} epochs (cf. Figure~S5 and Section~S5 of the Supplementary Material).

For evaluation, we applied the trained networks to our test datasets (see Table~\ref{tab:dataset}). %
We ran Darknet on the test images and stored corresponding detections (i.e., bounding-box locations and the confidence score of the network). %
Predictions for one image were computed in \SI{30}{\milli\second} in our implementation. 
Detections with a confidence score below \SI{.5}{\percent} were discarded. %
The test-time augmentation (including NMS) and AP, TP, and FP scores (as reported in Tables~\ref{tab:results_AOS} and \ref{tab:results_singleImg}) were computed in Matlab. %
For NMS and for AP, TP and FP scores we used an IoU threshold of \SI{25}{\percent}. %
To account for the AABB clipping and to avoid false defections at the image borders, we discarded detection results for which the AABB's center was too close to the image border. %
Since the median bounding box size in the training set was \SI{35}{\pixel}, we used a distance threshold of \SI{5}{\pixel} (half of \SI{35}{\pixel}$\times$\SI{25}{\percent}, rounded up). %
Note that this only affects single-image detection results, as our integral images require no AABB clipping.
For integral images, we obtained one detection result per scene (see Table~\ref{tab:results_AOS} and Figure~\ref{fig:results_AOS}).
For single images, we averaged detection results over all single images per scene (see Table~\ref{tab:results_singleImg}). Figure~\ref{fig:results_singleImg} shows a representative selection of single-image classification results. Precision and recall plots are shown in Figure~S6 and discussed in Section S6 of the Supplementary Material.

\subsection*{Ethical approval.}
The ethics committee of the Upper Austrian government approved the study, and participants provided written informed consent.

\section*{Data availability}  \label{sec:data}
The data collected in experiments with users can be downloaded from Zenodo\cite{OurData}, and includes labels and augmented images for training, validation, and testing, configuration files, trained network weights, and results. 

\section*{Code availability} \label{sec:code}
Code to compute Tables~\ref{tab:results_AOS} and \ref{tab:results_singleImg} is provided with the dataset.\cite{OurData} %
Further code that supports the findings of this study is available from the corresponding author upon reasonable request. %

\bibliography{references}
\section*{Corresponding author}
Communication and requests for material should be addressed to Oliver Bimber (email:~\href{mailto:oliver.bimber@jku.at}{oliver.bimber@jku.at}; orcid:~\href{https://orcid.org/0000-0001-9009-7827}{0000-0001-9009-7827}).

\section*{Acknowledgements}
This research was funded by the Austrian Science Fund (FWF) under grant number P 32185-NBL, and by the State of Upper Austria and the Austrian Federal Ministry of Education, Science and Research via the LIT -- Linz Institute of Technology under grant number LIT-2019-8-SEE-114. 
\section*{Author contributions statement}
D.S. and O.B. conceived and designed the experiments. %
D.S. and I.K. performed the experiments. %
D.S. and O.B. analyzed the data. %
D.S. and I.K. contributed materials/analysis tools. %
D.S. and O.B. wrote the paper. %
  		
\section*{Competing Interests}
The authors declare that they have no competing interests.

\end{document}


\flushbottom
\maketitle
\thispagestyle{empty}

\section{Computing AOS Integral Images} \label{sup:integral}
\begin{figure}[!b]
\centering
\includegraphics[width=\columnwidth]{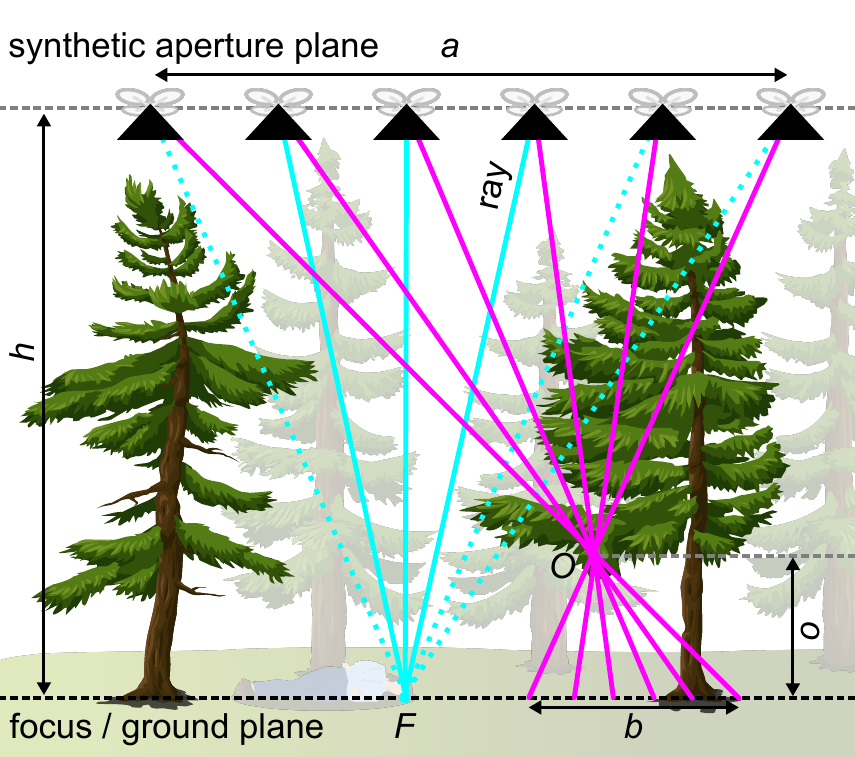} %
\caption{Schematic drawing of AOS' integration principle. Multiple images at the synthetic-aperture-plane of size $a$ and altitude $h$ above ground level are recorded and integrated as a 4D light field. %
Exemplary rays are shown for a target point $F$ (cyan) on the synthetic focal plane (aligned with the ground surface), and an out-of-focus occluder $O$ (magenta). While the rays of $F$ form a point on the synthetic focal plane, the projection of out-of-focus rays form an area of size $b$ (cf. \autoref{eq:blur}). %
Dotted in-focus rays indicate occlusion (e.g., a $D=50\%$ occlusion by vegetation). %
}%
\label{fig:integral} 
\end{figure}

In this section we briefly revisit the principles of Airborne Optical Sectioning  (AOS)\cite{Kurmi2018,Bimberi2019IEEECGA,Kurmi2019TAOS,Kurmi2019Sensor,Schedl2020SRep}.
The theoretical basis for AOS are unstructured light fields, which represent a 4D subset of the plenoptic function. The interested reader is referred to surveys\cite{Wetzstein2011survey,Wu2017survey} which cover the topic thoroughly. %
For AOS, thermal radiation (recorded by a thermal camera) is described by rays within an occlusion volume (forest), as show in \autoref{fig:integral}. We parameterize these rays by their intersection with two parallel planes: the synthetic aperture plane (2D directional coordinate at the drone's altitude above the ground $h$) and the synthetic focal plane (2D spatial coordinate on the ground). %
In practise, the drone's positions are not precisely on the synthetic aperture plane and the camera's orientation is not stable in flight.
Thus, precise 3D pose estimation (extrinsics) and camera calibration (intrinsics) are necessary (see Methods section of the main article). Once the intrinsic and extrinsic parameters are estimated, a single sensor reading (i.e., a thermal pixel) can be parmeterized as a single ray within the 4D light field, resulting in approximately \SI{94}{\mega\nothing} rays in total for each recorded scene.

For integration, rays that intersect at a given spatial coordinate (i.e., varying directional coordinates on the synthetic aperture plane and similar spatial coordinates on the synthetic focal plane) are averaged (cf. \autoref{fig:integral} in-focus point $F$). The result of integrating all rays across all spatial coordinates on the synthetic focal plane is a focused (on the ground) integral image. %
Setting the focus parameters (i.e., aligning the synthetic focal plane with the unknown ground surface) is done by automatic optimization\cite{Kurmi2020} (see Results and Methods sections of the main article). %
Note, that changing the focus parameters requires a reparameterization\cite{Isaksen2000} of the 4D ray space, as the spatial coordinates are changing. 

In case of occlusion, rays will be $D$ likely to contain signals of occluders, where $D$ is the density of occluders\cite{Kurmi2019Sensor}. Occluded rays reduce the contrast (cf. Figure~3) and form a scattered footprint in the integral image (cf. \autoref{fig:integral} point $O$). 
While occluders are blurred and deemphasized, targets on the focused ground (such as humans) can be made clearly visible. %
The footprint size $b$ of a hypothetical infinitely-small out-of-focus point can be expressed with the intercept theorem as %
\begin{equation}\label{eq:blur}%
    b = \frac{o a}{h-o} \text{ ,}%
\end{equation}%
where $o$ is the altitude (above ground level) of the occluder, $a$ is the synthetic aperture size, and $h$ is the height of the synthetic aperture plane above ground level. %
This means, for example, that an occluder at $o=\SI{2}{\m}$ above ground has a footprint $b=\SI{1.8}{\m}$, in our experiments ($a=\SI{30}{\m}$; $h=\SI{35}{\m}$). %
By considering the occluder size $w$, \autoref{eq:blur} extends to %
\begin{equation}\label{eq:blurW}%
    b' = b + \frac{w h}{h-o} = \frac{o a + w h}{h-o} \text{ ,}%
\end{equation}%
where ${(w h)}/{(h-o)}$ is the occluder's projection onto the focal (ground) plane.

\section{Using Existing Image Datasets} \label{sup:flir}
\begin{figure*}[!th]
\centering
\includegraphics[width=\linewidth]{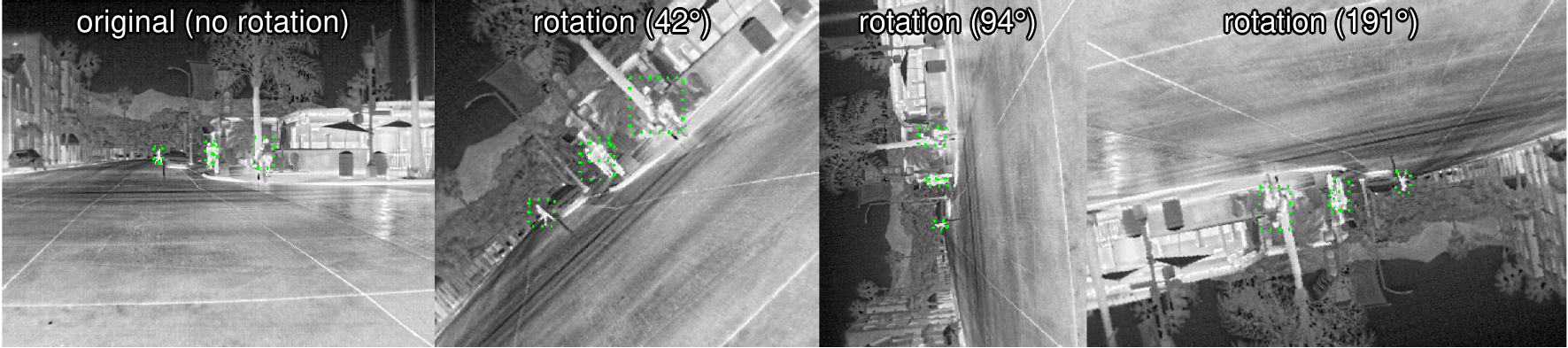}
\caption{An image from the ADAS dataset used for experiments in \autoref{sup:flir}. Exemplary rotations (and crops to avoid the introduction of borders), used for augmentation, are shown. Person labels are indicated by dashed green rectangles. Note that the axis-aligned bounding boxes change size due to rotation. %
}%
\label{fig:flir} 
\end{figure*}  
In this section, we apply the FLIR advanced driver-assistance systems (ADAS)\cite{FLIRadas} dataset (used for autonomous driving) for training and evaluate its performance when applied to our single image recordings. %
The dataset consists of \num{14452} annotated thermal images with \num{50116} person labels and was recorded with a thermal sensor that is similar to ours (\SI{14}{\bit}; \SI{640 x 512}{\pixel}). %
It contains mainly upright standing persons, thus we optionally apply \num{10} random rotations, in a \num{360} degree range, on the images and on the labels (cf. \autoref{fig:flir}). %
Note that we did not apply the optional adaptive histogram equalization (AHE) on the training set for this experiment, as it did not improve single-image results (cf. Table~3 of the main manuscript). %
We use \SI{90}{\percent} of the thermal images for training and \SI{10}{\percent} for validation and perform training as discussed in the Methods section of the main article. 
The training weights with the highest AP score, when applied to the validation dataset, are obtained after \num{2400} iterations or \num{1} epoch. %
Evaluation results without and with AHE applied to the test images are shown in \autoref{tab:flir}. %
Detection performance on our single images is inferior (the best overall AP scores are below \SI{0.1}{\percent}), thus indicating that available pedestrian datasets cannot be applied for aerial imaging. %
The labels in the ADAS datasets contain standing, walking, running and biking persons, while our aerial images show only lying humans. %
Nevertheless, we believe that the main reason for the poor results is caused by the difference in environments: the ADAS dataset contains persons in hot urban environments, while our recordings show comparatively cool forests with only a few heat spots. %
Despite the fact that other augmentation techniques (e.g., inverting the temperature scale) may improve detection performance for existing datasets, we believe that our specialized dataset will continue to outperform others. Furthermore, for AOS we rely on known camera poses which are not available for other existing datasets.
\begin{table}[htbp]
\centering %
\caption{Single-image person detection results with a network trained on a pedestrian dataset. Average precision scores (AP), ground truth (GT), true positives (TP), and false positives (FP) for each scene. We compare cases where adaptive histogram equalization (AHE) is applied and not applied to the test set. The scores are clearly worse when compared to Table~3 in the main article.}%
\begingroup %
\setlength{\tabcolsep}{5pt} %
\begin{tabular}{r@{ }l|rrrrrr}
      &       & \multicolumn{3}{c|}{\textbf{no AHE}} & \multicolumn{3}{c}{\textbf{AHE test set only}} \\
\textbf{ID} & \textbf{(GT)} & \multicolumn{1}{c}{\textbf{AP}} & \multicolumn{1}{l}{\textbf{FP}} & \multicolumn{1}{l|}{\textbf{TP}} & \multicolumn{1}{c}{\textbf{AP}} & \multicolumn{1}{l}{\textbf{FP}} & \multicolumn{1}{l}{\textbf{TP}} \bigstrut\\
\hline
F0    & (2.6) & 0.01\% & 0.10  & \multicolumn{1}{r|}{0.00} & \textbf{0.66\%} & 16.14 & \textbf{0.38} \bigstrut[t]\\
F1    & (7.2) & 0.00\% & 0.29  & \multicolumn{1}{r|}{0.00} & \textbf{0.04\%} & 2.06  & \textbf{0.06} \\
F2    & (8.5) & 0.00\% & 0.16  & \multicolumn{1}{r|}{0.00} & \textbf{0.00\%} & 3.49  & \textbf{0.02} \\
F3    & (4.4) & 0.04\% & 0.09  & \multicolumn{1}{r|}{0.01} & \textbf{0.29\%} & 8.08  & \textbf{0.24} \\
F4    & (2.6) & 0.00\% & 0.07  & \multicolumn{1}{r|}{0.00} & \textbf{0.02\%} & 7.22  & \textbf{0.05} \\
F5    & (5.8) & 0.00\% & 0.00  & \multicolumn{1}{r|}{0.00} & \textbf{0.06\%} & 0.01  & \textbf{0.01} \\
F6    & (5.7) & 0.00\% & 0.01  & \multicolumn{1}{r|}{0.00} & 0.00\% & 0.02  & 0.00 \\
F7    & (2.0) & 0.00\% & 0.00  & \multicolumn{1}{r|}{0.00} & \textbf{1.61\%} & 0.00  & \textbf{0.03} \\
F9    & (0.0) & \multicolumn{1}{c}{n/a} & 0.02  & \multicolumn{1}{r|}{0.00} & \multicolumn{1}{c}{n/a} & 1.54  & 0.00 \\
F11   & (0.0) & \multicolumn{1}{c}{n/a} & 0.62  & \multicolumn{1}{r|}{0.00} & \multicolumn{1}{c}{n/a} & 8.40  & 0.00 \bigstrut[b]\\
\hline
\textbf{avg} & (3.6) & 0.00\% & 0.15  & 0.00  & \textbf{0.09\%} & 5.80  & \textbf{0.09} \bigstrut[t]\\
\end{tabular}%
\endgroup
\label{tab:flir}%
\end{table}

\section{Test and Training Sites} \label{sup:dataset}
\begin{figure*}
\centering
\includegraphics[width=\linewidth]{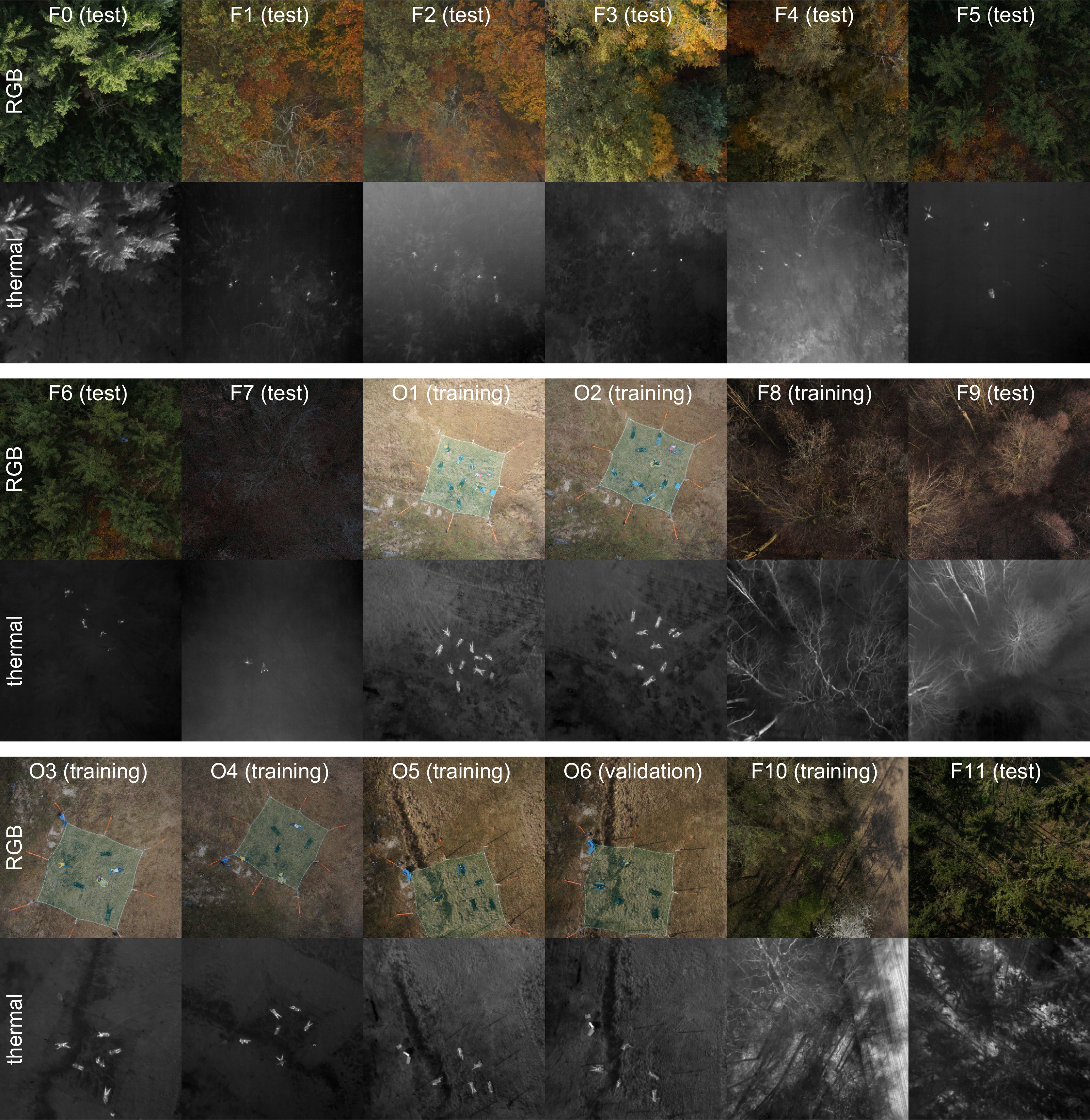}
\caption{Exemplary RGB and thermal images of the 18 flights that were used to create the training, validation, and test sets. Further details can be found in Table~1 in the main article.%
}%
\label{fig:dataset} 
\end{figure*}

\autoref{fig:dataset} shows a representative selection of single RGB and thermal images of all \num{18} flights at \num{6} different sites over \num{10} different days used for the training, validation and test sets. Note that the RGB images are only used for pose estimation (see Methods section in the main article). %
\begin{figure*}[!hbt]
\centering
\includegraphics[width=\linewidth]{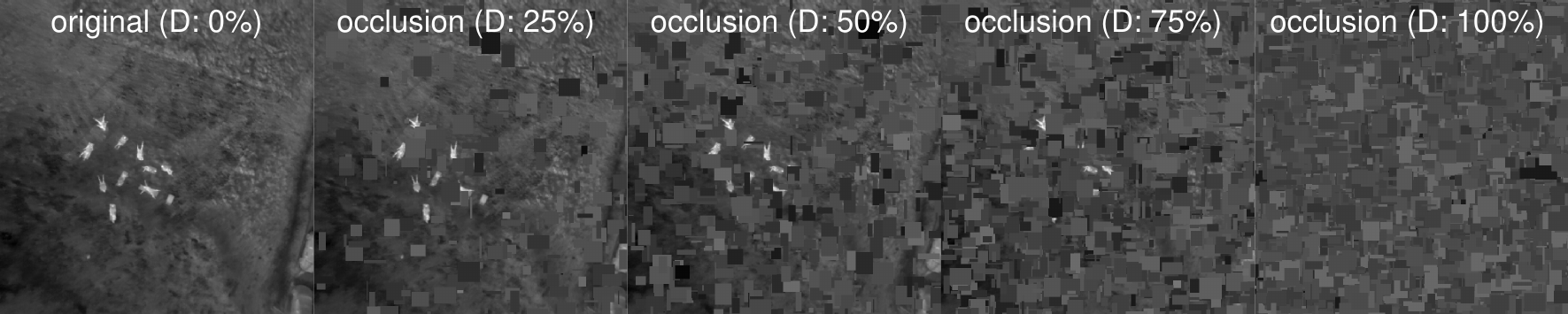}
\caption{Simulated occlusion for various densities $D$ applied to a single image of the open field training data set. %
}%
\label{fig:occlusion} 
\end{figure*}  
Scenes labeled with the letter \textbf{O} are recorded above our open field training area. %
The green safety net is clearly visible in the RGB images, but not resolved in the thermal recordings due to the thin strings (\SI{4.75}{\milli\m}) of the net and the comparatively low resolution (\SI{640 x 512}{\pixel} raw; \SI{512 x 512}{\pixel} rectified and cropped) of the thermal camera. %
Scenes labeled with \textbf{F} are recorded above forest. The RGB images illustrate the density and vegetation of the different sites. The complete dataset, and flight logs are available open access\cite{OurData}.

\section{Augmenting Simulated Occlusion} \label{sup:occlusion}
In this section we report on initial experiments that augment simulated occlusion to investigate if this can improve the single image detection rate. 
We apply an augmentation strategy that is similar to random erasing\cite{Zhong2020erasing} for modelling a Bernoulli distributed occlusion pattern \cite{Kurmi2019Sensor}. %
A random density $D$ (\SIrange{0}{100}{\percent}) defines the probability of occlusion in an image (cf. Figure~\ref{fig:occlusion}). We model occluders as axis-aligned rectangles of random width and height (\SIrange{1}{35}{\pixel} i.e., the average bounding box size in the training set). Thermal values for the occluding rectangles are random samples from non-labeled (i.e., background) regions of the original single images. %
We apply occlusion augmentation for (initially every occlusion free) image in the training and validation dataset and add it to the corresponding set prior to training. %
The best training weights are retrieved after \num{2800} iterations or \num{4} epochs. %
Note, that we did not apply the optional adaptive histogram equalization (AHE) to the training set for this experiment, as it did not improve single-image results (cf. Table~3 of the main manuscript). %
Evaluation results (shown in \autoref{tab:occlusion}), indicate that the detection rate with single images is not improved by augmenting simulated occlusion.

\begin{table}[htbp]
\centering %
\caption{Single-image person detection results with simulated occlusion augmentation. Average precision scores (AP), ground truth (GT), true positives (TP), and false positives (FP) for each scene. Augmentation with AHE: not applied, and applied to the test set only. Note, that the scores did not improve, when compared to Table~3 in the main article.}%
\begin{tabular}{r@{ }l|rrrrrr}
      &       & \multicolumn{3}{c|}{\textbf{no AHE}} & \multicolumn{3}{c}{\textbf{AHE test set only}} \\
\textbf{ID} & \textbf{(GT)} & \multicolumn{1}{c}{\textbf{AP}} & \multicolumn{1}{l}{\textbf{FP}} & \multicolumn{1}{l|}{\textbf{TP}} & \multicolumn{1}{c}{\textbf{AP}} & \multicolumn{1}{l}{\textbf{FP}} & \multicolumn{1}{l}{\textbf{TP}} \bigstrut[b]\\
\hline
F0    & (2.6) & 1.9\% & 7.3   & \multicolumn{1}{r|}{0.5} & \textbf{2.9\%} & 7.5   & \textbf{0.6} \bigstrut[t]\\
F1    & (7.2) & 41.9\% & 1.6   & \multicolumn{1}{r|}{3.4} & \textbf{44.8\%} & 1.8   & \textbf{3.6} \\
F2    & (8.5) & 20.3\% & 1.4   & \multicolumn{1}{r|}{2.2} & \textbf{22.8\%} & 1.5   & \textbf{2.4} \\
F3    & (4.4) & 7.0\% & 0.3   & \multicolumn{1}{r|}{0.4} & \textbf{10.2\%} & 0.4   & \textbf{0.6} \\
F4    & (2.6) & 3.0\% & 0.7   & \multicolumn{1}{r|}{0.2} & \textbf{4.2\%} & 0.7   & \textbf{0.3} \\
F5    & (5.8) & 47.1\% & 0.4   & \multicolumn{1}{r|}{2.8} & \textbf{53.0\%} & 0.4   & \textbf{3.1} \\
F6    & (5.7) & 58.3\% & 0.4   & \multicolumn{1}{r|}{3.4} & \textbf{60.6\%} & 0.4   & \textbf{3.5} \\
F7    & (2.0) & 25.3\% & 0.0   & \multicolumn{1}{r|}{0.5} & \textbf{87.1\%} & 0.0   & \textbf{1.7} \\
F9    & (0.0) & \multicolumn{1}{c}{n/a} & 0.9   & \multicolumn{1}{r|}{0.0} & \multicolumn{1}{c}{n/a} & 0.9   & 0.0 \\
F11   & (0.0) & \multicolumn{1}{c}{n/a} & 14.4  & \multicolumn{1}{r|}{0.0} & \multicolumn{1}{c}{n/a} & 14.4  & 0.0 \bigstrut[b]\\
\hline
\textbf{avg} & (3.6) & 13.5\% & 3.5   & 1.2   & \textbf{18.38\%} & 3.5   & \textbf{1.3} \bigstrut[t]\\
\end{tabular}%
\label{tab:occlusion}%
\end{table}%

\section{Training Weights} \label{sup:training}
\begin{figure*}
\centering
\includegraphics[width=\linewidth]{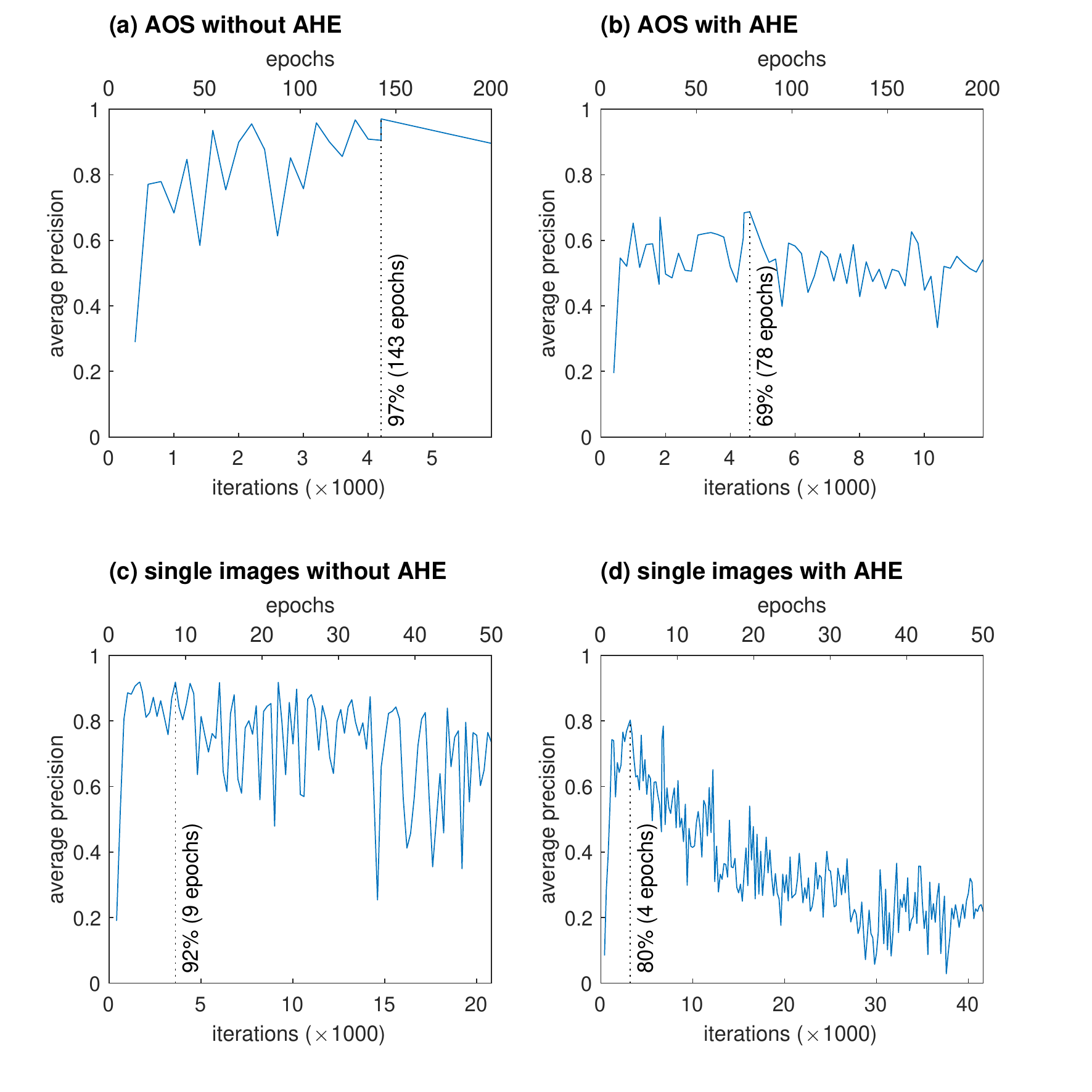}
\caption{Average precision scores (IoU \SI{50}{\percent}) on the validation dataset during training for the AOS integral images (a,b) and the single images (c,d), without (a,c) and with (b,d) the optional AHE contrast augmentation. Network weights are evaluated after every 200 training iterations and the weights with maximum AP are used for the evaluation. The bottom x-axis denotes iterations (in thousands) and the top x-axis illustrates the number of epochs. Note that the number of iterations for one epoch is depending on the dataset size (i.e., the number of training images).%
}%
\label{fig:training} 
\end{figure*}

Training weights are stored after every \num{200} batch iterations and the weight with the highest AP scores (with IoU \SI{50}{\percent} using Darknet) on the validation datasets are used as final weight. %
Note, that for the validation and training sets the same training augmentations were applied.
\autoref{fig:training} plots AP scores on the validation dataset during training of the AOS integral images and the single images. %
Highest AP scores are achieved after \num{4200} and \num{4600} iterations, and \num{143} and \num{78} epochs for AOS without and with AHE training augmentation, respectively. %
For single images, the highest AP scores are achieved after \num{3600} and \num{3200} iterations, and \num{9} and \num{4} epochs without and with AHE training augmentation. %
Note, that the number of iterations for one epoch is depending on the number of training images (i.e., more training images requires a higher number of iterations for one epoch). No test augmentations were applied on the validation datasets. 

\section{Precision Recall Plots} \label{sup:precision_recall}
\begin{figure}
\centering
\includegraphics[width=\columnwidth]{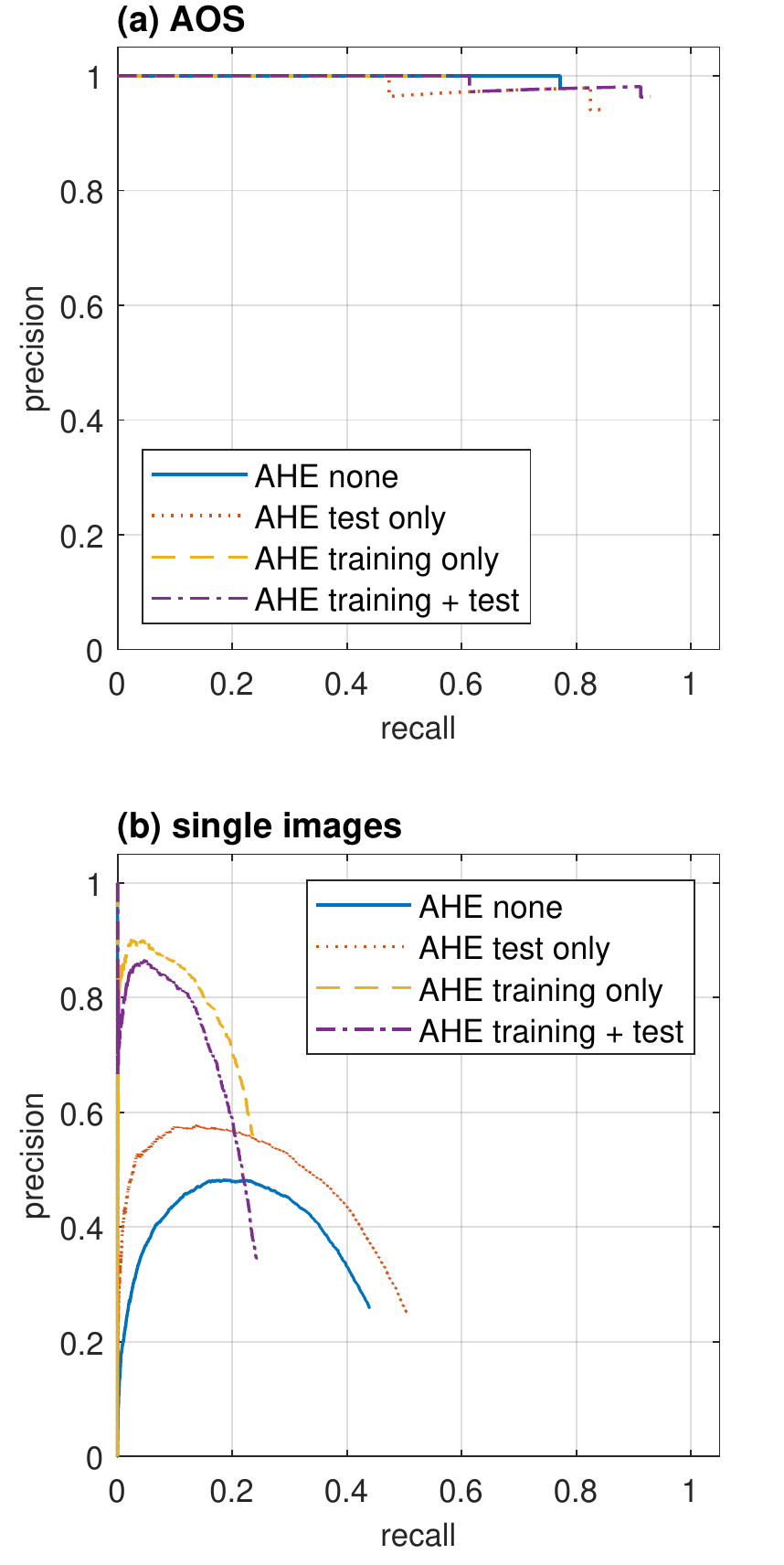}
\caption{Precision-recall plots for the results presented in the main article (Table~2, (a) and 3, (b)). The blue solid lines indicate results without the optional AHE contrast augmentation. The red dotted lines show results with AHE applied to the test set only, while the orange dashed lines show results with AHE applied to the training set only. The violet, dot and dashed lines show results when AHE is applied to the training and the test set. %
}%
\label{fig:precision_recall} 
\end{figure}

Precision-recall plots for the evaluation results presented in the main article (Table~2 and 3) are shown in \autoref{fig:precision_recall}. We used Matlab to compute the scores and plot the curves. The precision-recall curves are the basis for AP score computations. %
For AOS (cf. \autoref{fig:precision_recall}(a)) the precision scores are only slightly below \num{1.0} (due to the low number of FPs), while still achieving maximum recall scores of \num{.77}, \num{.84}, \num{.58}, and \num{.93} for no AHE augmentation, test-time AHE, AHE applied to training only, and AHE applied to training and test, respectively. %
Precision and recall scores for single images (cf. \autoref{fig:precision_recall}(b)) are clearly worse, when compared to AOS results. For all four cases, recall is never higher than \num{.5}. Ignoring the initial precision scores of \num{1}, the highest single-image precision/recall scores are \num{.48}/\num{.44}, \num{.58}/\num{.50}, \num{.9}/\num{.24}, and \num{.87}/\num{.24} for no AHE augmentation, and AHE applied to test only, training only, and to the training and test set, respectively. %
The network which is trained without AHE augmentation (AHE none and AHE test only) achieves higher recall but lower precision scores (due to a high number of FPs) when compared to the network that is trained with AHE augmentation on the training set (AHE training only and AHE trainind + test).

\bibliography{references}